\documentclass[11pt,a4paper]{article}
\usepackage{authblk}
\usepackage[hyperref]{emnlp-ijcnlp-2019}
\usepackage{times}
\usepackage{latexsym}
\usepackage{caption}
\usepackage{enumitem}

\usepackage{url}

\usepackage{graphicx} 
\graphicspath{ {images/} }

\usepackage{comment} 

\usepackage{amssymb} 
\usepackage{amsmath} 

\usepackage{booktabs} 

\usepackage{paralist} 

\PassOptionsToPackage{hyphens}{url}\usepackage{hyperref} 

\newcommand{\tberg}[1]{\textcolor{blue}{\bf \small [ #1 --TB]}}

\aclfinalcopy 

\title{A Deep Factorization of Style and Structure in Fonts}

\author[$1$]{Nikita Srivatsan}
\author[$2$]{Jonathan T. Barron}
\author[$3$]{Dan Klein}
\author[$4$]{Taylor Berg-Kirkpatrick}
\affil[$1$]{Language Technologies Institute\\ Carnegie Mellon University\\ \texttt{nsrivats@cmu.edu}}
\affil[$2$]{Google Research\\ \texttt{barron@google.com}}
\affil[$3$]{Computer Science Division\\ University of California, Berkeley\\ \texttt{klein@cs.berkeley.edu}}
\affil[$4$]{Computer Science and Engineering\\ University of California, San Diego\\ \texttt{tberg@eng.ucsd.edu}}

\date{}

\begin{document}

\twocolumn[{%
\renewcommand\twocolumn[1][]{#1}%
\maketitle
\begin{center}
\includegraphics[width=\linewidth]{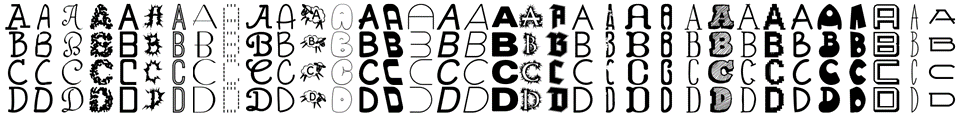}
\captionof{figure}{Example fonts from the Capitals64 dataset. The task of font reconstruction involves generating missing glyphs from partially observed novel fonts.}\label{fig:font-matrix}
\vspace{1.0em}
\end{center}}]


\begin{abstract}

We propose a deep factorization model for typographic analysis that disentangles \emph{content} from \emph{style}. Specifically, a variational inference procedure factors each training glyph into the combination of a character-specific content embedding and a latent font-specific style variable. The underlying generative model combines these factors through an asymmetric transpose convolutional process to generate the image of the glyph itself. When trained on corpora of fonts, our model learns a manifold over font styles that can be used to analyze or reconstruct new, unseen fonts. On the task of reconstructing missing glyphs from an unknown font given only a small number of observations, our model outperforms both a strong nearest neighbors baseline and a state-of-the-art discriminative model from prior work.

\end{abstract}

\section{Introduction}

One of the most visible attributes of digital language data is its typography. A font makes use of unique stylistic features in a visually consistent manner across a broad set of characters while preserving the structure of each underlying form in order to be human readable -- as shown in Figure~\ref{fig:font-matrix}. Modeling these stylistic attributes and how they compose with underlying character structure could aid typographic analysis and even allow for automatic generation of novel fonts. Further, the variability of these stylistic features presents a challenge for optical character recognition systems, which typically presume a library of known fonts. In the case of historical document recognition, for example, this problem is more pronounced due to the wide range of lost, ancestral fonts present in such data~\citep{ocular1, ocular2}. Models that capture this wide stylistic variation of glyph images may eventually be useful for improving optical character recognition on unknown fonts.   

In this work we present a probabilistic latent variable model capable of disentangling stylistic features of fonts from the underlying structure of each character. Our model represents the style of each font as a vector-valued latent variable, and parameterizes the structure of each character as a learned embedding. Critically, each style latent variable is shared by all characters within a font, while character embeddings are shared by characters of the same type across all fonts. Thus, our approach is related to a long literature on using tensor factorization as a method for disentangling style and content \cite{freeman1997learning,tenenbaum2000separating,vasilescu2002multilinear,tang2013tensor} and to recent deep tensor factorization techniques~\cite{deepmatrix}.

Inspired by neural methods' ability to disentangle loosely coupled phenomena in other domains, including both language and vision~\citep{hu2017controllable,yang2017improved,gatys2016image,zhu2017unpaired}, we parameterize the distribution that combines style and structure in order to generate glyph images as a transpose convolutional neural decoder~\cite{dumoulin2016guide}. Further, the decoder is fed character embeddings early on in the process, while the font latent variables directly parameterize the convolution filters. This architecture biases the model to capture the asymmetric process by which structure and style combine to produce an observed glyph.

We evaluate our learned representations on the task of {\em font reconstruction}. After being trained on a set of observed fonts, the system reconstructs missing glyphs in a set of previously unseen fonts, conditioned on a small observed subset of glyph images. Under our generative model, font reconstruction can be performed via posterior inference. Since the posterior is intractable, we demonstrate how a variational inference procedure can be used to perform both learning and accurate font reconstruction. In experiments, we find that our proposed latent variable model is able to substantially outperform both a strong nearest-neighbors baseline as well as a state-of-the-art discriminative system on a standard dataset for font reconstruction. Further, in qualitative analysis, we demonstrate how the learned latent space can be used to interpolate between fonts, hinting at the practicality of more creative applications. 



\section{Related Work}

Discussion of computational style and content separation in fonts dates at least as far back as the writings of~\citet{hofstadter1983metamagical,hofstadter1995fluid}. Some prior work has tackled this problem through the use of bilinear factorization models~\cite{freeman1997learning,tenenbaum2000separating}, while others have used discriminative neural models \cite{zhang2018separating,zhang2020unified} and adversarial training techniques~\cite{GAN}. In contrast, we propose a deep probabilistic approach that combines aspects of both these lines of past work. Further, while some prior approaches to modeling fonts rely on stroke or topological representations of observed glyphs~\cite{campbell2014learning,phan2015flexyfont,suveeranont2010example}, ours directly models pixel values in rasterized glyph representations and allows us to more easily generalize to fonts with variable glyph topologies.

Finally, while we focus our evaluation on font reconstruction, our approach has an important relationship with style transfer -- a framing made explicit by~\citet{zhang2018separating,zhang2020unified} -- as the goal of our analysis is to learn a smooth manifold of font styles that allows for stylistic inference given a small sample of glyphs. However, many other style transfer tasks in the language domain~\cite{shen2017style} suffer from ambiguity surrounding the underlying division between style and semantic content. By contrast, in this setting the distinction is clearly defined, with content (i.e.\ the character) observed as a categorical label denoting the coarse overall shape of a glyph, and style (i.e.\ the font) explaining lower-level visual features such as boldness, texture, and serifs. The modeling approach taken here might inform work on more complex domains where the division is less clear.

\section{Font Reconstruction}

\begin{figure*}
\centering
\includegraphics[width=0.82\linewidth]{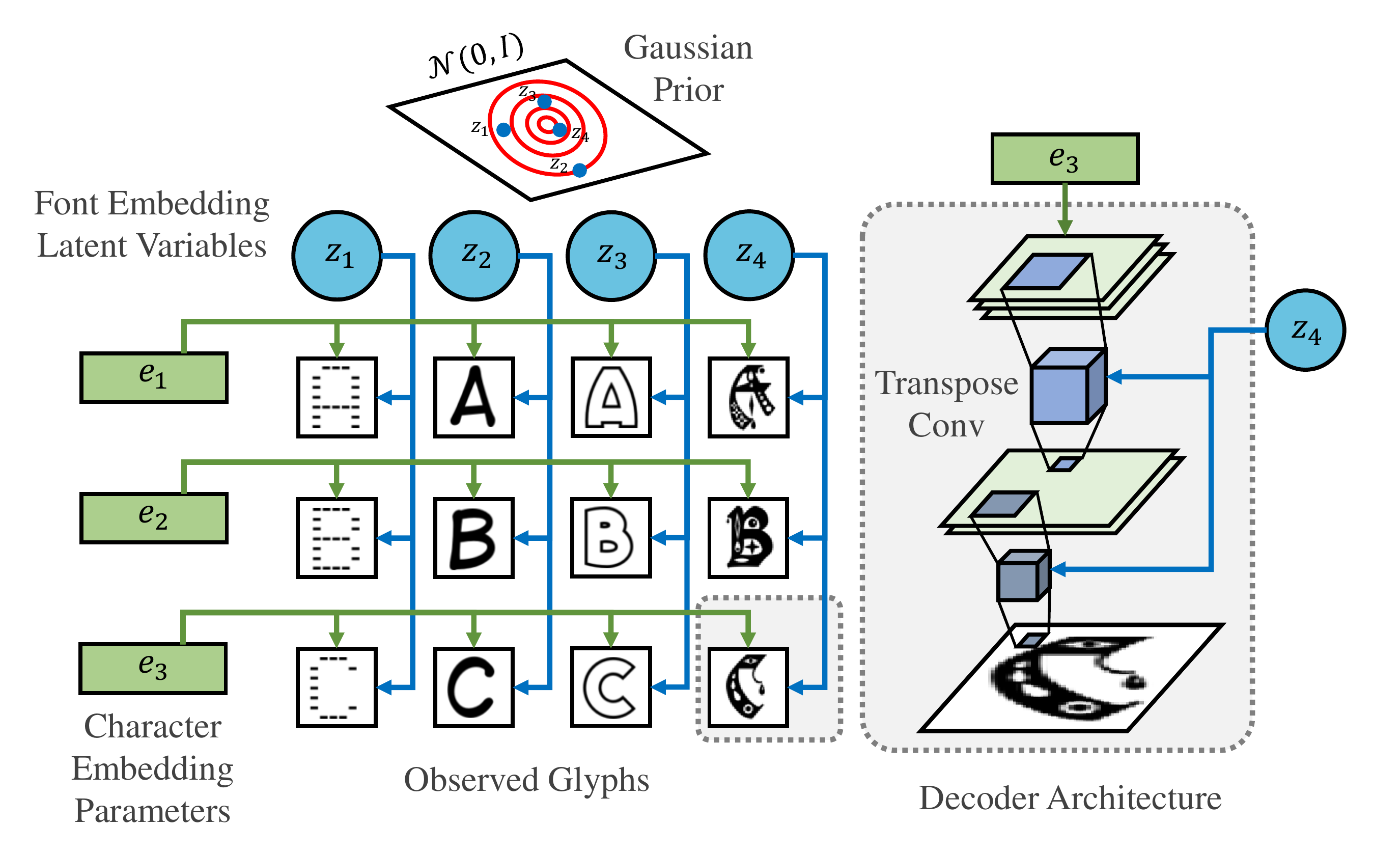}
\caption{Depiction of the generative process of our model. Each observed glyph image is generated conditioned on the latent variable of the corresponding font and the embedding parameter of the corresponding character type. For a more detailed description of the decoder architecture and hyperparameters, see Appendix~\ref{sec:appendix}.}
\label{fig:model}
\end{figure*}

We can view a collection of fonts as a matrix, $X$, where each column corresponds to a particular character type, and each row corresponds to a specific font. Each entry in the matrix, $x_{ij}$, is an image of the glyph for character $i$ in the style of a font $j$, which we observe as a $64\times64$ grayscale image as shown in Figure~\ref{fig:font-matrix}. In a real world setting, the equivalent matrix would naturally have missing entries wherever the encoding for a character type in a font is undefined. In general, not all fonts contain renderings of all possible character types; many will only support one particular language or alphabet and leave out uncommon symbols. Further, for many commercial applications, only the small subset of characters that appears in a specific advertisement or promotional message will have been designed by the artist -- the majority of glyphs are missing. As a result, we may wish to have models that can infer these missing glyphs, a task referred to as font reconstruction. 

Following recent prior work~\cite{GAN}, we define the task setup as follows: During training we have access to a large collection of observed fonts for the complete character set.  At test time we are required to predict the missing glyph images in a collection of {\em previously unseen} fonts with the same character set. Each test font will contain observable glyph images for a small randomized subset of the character set. Based on the style of this subset, the model must reconstruct glyphs for the rest of the character set. 

Font reconstruction can be thought of as a form of matrix completion; given various observations in both a particular row and column, we wish to reconstruct the element at their intersection. Alternatively we can view it as a few-shot style transfer task, in that we want to apply the characteristic attributes of a new font (e.g. serifs, italicization, drop-shadow) to a letter using a small number of examples to infer those attributes. Past work on font reconstruction has focused on discriminative techniques. For example~\citet{GAN} used an adversarial network to directly predict held out glyphs conditioned on observed glyphs. By contrast, we propose a generative approach using a deep latent variable model. Under our approach fonts are generated based on an unobserved style embedding, which we can perform inference over given any number of observations.

\section{Model}

Figure~\ref{fig:model} depicts our model's generative process. Given a collection of images of glyphs consisting of $I$ character types across $J$ fonts, our model hypothesizes a separation of character-specific structural attributes and font-specific stylistic attributes into two different representations. Since all characters are observed in at least one font, each character type is represented as an embedding vector which is part of the model's parameterization. In contrast, only a subset of fonts is observed during training and our model will be expected to generalize to reconstructing unseen fonts at test time. Thus, our representation of each font is treated as a vector-valued latent variable rather than a deterministic embedding. 

More specifically, for each font in the collection, a font embedding variable, $z_j \in \mathbb{R}^{k}$, is sampled from a fixed multivariate Gaussian prior, $p(z_j) = \mathcal{N}(0,I_{k})$. Next, each glyph image, $x_{ij}$, is generated independently, conditioned on the corresponding font variable, $z_j$, and a character-specific parameter vector, $e_i \in \mathbb{R}^{k}$, which we refer to as a character embedding. Thus, glyphs of the same character type share a character embedding, while glyphs of the same font share a font variable. A corpus of $I * J$ glyphs is modeled with only $I$ character embeddings and $J$ font variables, as seen in the left half of Figure~\ref{fig:model}. This modeling approach can be thought of as a form of deep matrix factorization, where the content at any given cell is purely a function of the vector representations of the corresponding row and column.
We denote the full corpus of glyphs as a matrix $X = \left(\left(x_{11},...,x_{1J}\right),...,\left(x_{I1},...x_{IJ}\right)\right)$ and denote the corresponding character embeddings as $E = \left( e_1, ..., e_I \right)$ and font variables as $Z = \left( z_1, ..., z_J \right)$.

Under our model, the probability distribution over each image, conditioned on $z_j$ and $e_i$, is parameterized by a neural network, described in the next section and depicted in Figure~\ref{fig:model}. We denote this decoder distribution as $p(x_{ij} | z_j ; e_i,\phi)$, and let $\phi$ represent parameters, shared by all glyphs, that govern how font variables and character embeddings combine to produce glyph images. In contrast, the character embedding parameters, $e_i$, which feed into the decoder, are only shared by glyphs of the same character type. The font variables, $z_j$, are unobserved during training and will be inferred at test time. The joint probability under our model is given by:  
%
\begin{equation*}
\begin{split}
    p(X,Z;E,\phi) = \prod_{i,j} p(x_{ij} | z_j ; e_i,\phi) p(z_j)
\end{split}
\end{equation*}

\subsection{Decoder Architecture}

One way to encourage the model to learn disentangled representations of style and content is by choosing an architecture that introduces helpful inductive bias. For this domain, we can think of the character type as specifying the overall shape of the image, and the font style as influencing the finer details; we formulate our decoder with this difference in mind. We hypothesize that a glyph can be modeled in terms of a low-resolution character representation to which a complex operator specific to that font has been applied.

The success of transpose convolutional layers at generating realistic images suggests a natural way to apply this intuition. A transpose convolution\footnote{sometimes erroneously referred to as a ``deconvolution''} is a convolution performed on an undecimated input (i.e. with zeros inserted in between pixels in alternating rows and columns), resulting in an upscaled output. Transpose convolutional architectures generally start with a low resolution input which is passed through several such layers, iteratively increasing the resolution and decreasing the number of channels until the final output dimensions are reached. We note that the asymmetry between the coarse input and the convolutional filters closely aligns with the desired inductive biases, and therefore use this framework as a starting point for our architecture.

Broadly speaking, our architecture represents the underlying shape that defines the specific character type (but not the font) as coarse-grained information that therefore enters the transpose convolutional process early on. In contrast, the stylistic content that specifies attributes of the specific font (such as serifs, drop shadow, texture) is represented as finer-grained information that enters into the decoder at a later stage, by parameterizing filters, as shown in the right half of Figure~\ref{fig:model}. Specifically we form our decoder as follows: first the character embedding is projected to a low resolution matrix with a large number of channels. Following that, we apply several transpose convolutional layers which increase the resolution, and reduce the number of channels. Critically, the convolutional filter at each step is not a learned parameter of the model, but rather the output of a small multilayer perceptron whose input is the font latent variable $z$. Between these transpose convolutions, we insert vanilla convolutional layers to fine-tune following the increase in resolution.

Overall, the decoder consists of four blocks, where each block contains a transpose convolution, which upscales the previous layer and reduces the number of channels by a factor of two, followed by two convolutional layers. Each (transpose) convolution is followed by an instance norm and a ReLU activation. The convolution filters all have a kernel size of $5\times5$. The character embedding is reshaped via a two-layer MLP into a $8\times8\times256$ tensor before being fed into the decoder. The final $64\times64$ dimensional output layer is treated as a grid of parameters which defines the output distribution on pixels. We describe the specifics of this distribution in the next section.

\subsection{Projected Loss}

\begin{figure*}
\centering
\includegraphics[width=0.9\textwidth]{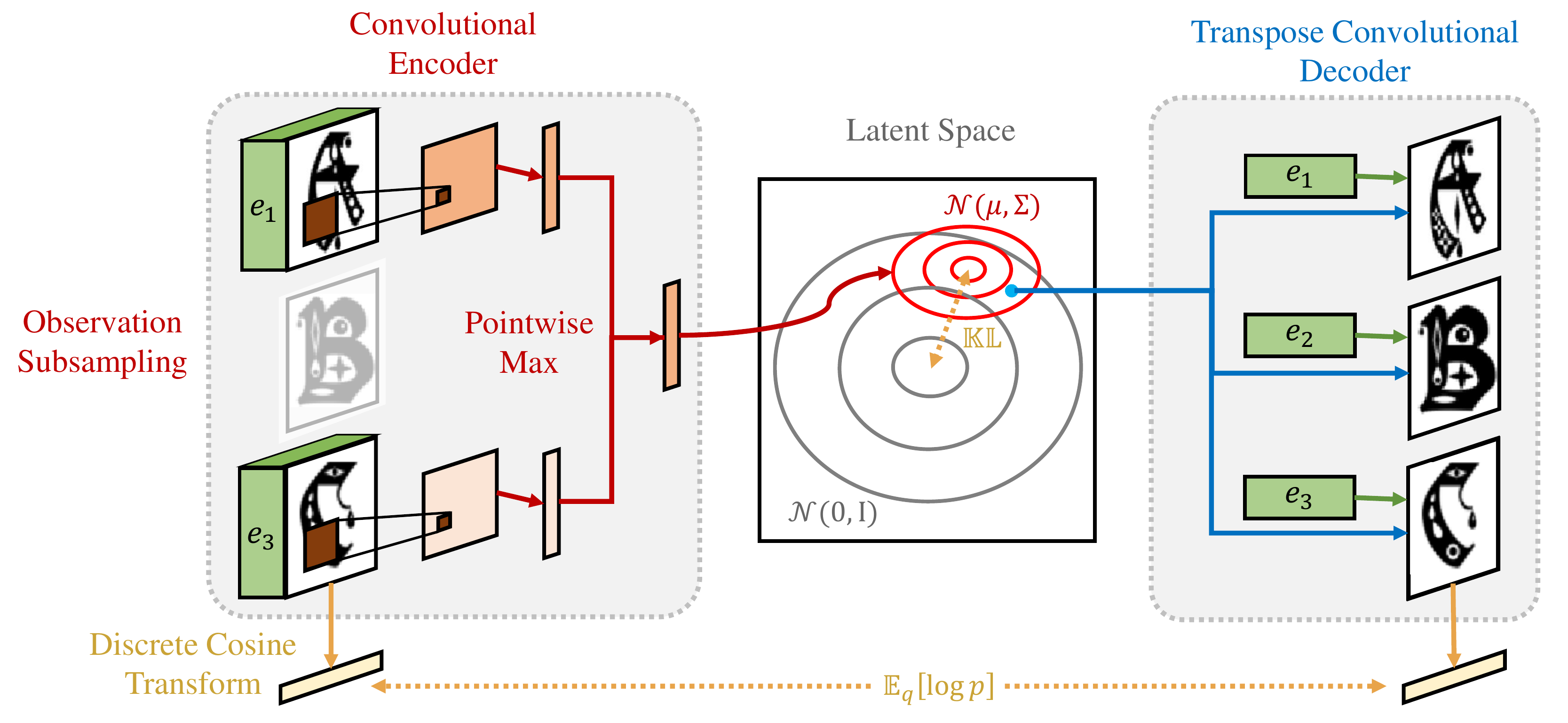}
\caption{Depiction of the computation graph of the amortized variational lower bound (for simplicity, only one font is shown). The encoder approximates the generative model's true posterior over the font style latent variables given the observations. It remains insensitive to the number of observations by pooling high-level features across glyphs. For a more specific description of the encoder architecture details, see Appendix~\ref{sec:appendix}.}
\label{fig:inference}
\end{figure*}

\newcommand{\postsample}{\hat{x_{ij}}}

The conventional approach for computing loss on image observations is to use an independent output distribution, typically a Gaussian, on each pixel's intensity. However, deliberate analysis of the statistics of natural images has shown that images are not well-described in terms of statistically independent pixels, but are instead better modeled in terms of edges~\cite{Field:87, Huang99}.
It has also been demonstrated that images of text have similar statistical distributions as natural images~\cite{Melmer13:Writing}. Following this insight, as our reconstruction loss we use a heavy-tailed (leptokurtotic) distribution placed on a transformed representation of the image, similar to the approach of \citet{BarronCVPR2019}. Modeling the statistics of font glyphs in this fashion results in sharper samples, while modeling independent pixels with a Gaussian distribution results in blurry, oversmoothed results.

More specifically, we adopt one of the strategies employed in \citet{BarronCVPR2019}, and transform image observations using the orthonormal variant of the 2-Dimensional Discrete Cosine Transform (2-D DCT-II) \cite{ahmed1974discrete}, which we denote as $f : \mathbb{R}^{64 \times 64} \rightarrow \mathbb{R}^{64 \times 64}$ for our $64 \times 64$ dimensional image observations. We transform both the observed glyph image and the corresponding grid or parameters produced by our decoder before computing the observation's likelihood. 

This procedure projects observed images onto a grid of orthogonal bases comprised of shifted and scaled 2-dimensional cosine functions.
Because the DCT-II is orthonormal, this transformation is volume-preserving, and so likelihoods computed in the projected space correspond to valid measurements in the original pixel domain.

Note that because the DCT-II is simply a rotation in our vector space, imposing a normal distribution in this transformed space should have little effect (ignoring the scaling induced by the diagonal of the covariance matrix of the Gaussian distribution) as Euclidean distance is preserved under rotations. For this reason we impose a heavy-tailed distribution in this transformed space, specifically a Cauchy distribution. This gives the following probability density function

\begin{equation*}
g(x; \hat{x}, \gamma)= {\frac {1}{\pi \gamma \,\left(1+\left({\frac {f(x)-f(\hat{x})}{\gamma }}\right)^{2}\right)}}
\end{equation*}

where $x$ is an observed glyph, $\hat{x}$ is the location parameter grid output by our decoder, and $\gamma$ is a hyperparameter which we set to $\gamma=0.001$.

The Cauchy distribution accurately captures the heavy-tailed structure of the edges in natural images. Intuitively, it models the fact that images tend to be mostly smooth, with a small amount of non-smooth variation in the form of edges.
Computing this heavy-tailed loss over the frequency decomposition provided by the DCT-II instead of the raw pixel values encourages the decoder to generate sharper images without needing either an adversarial discriminator or a vectorized representation of the characters during training.
Note that while our training loss is computed in DCT-II space, at test time we treat the raw grid of parameter outputs $\hat{x}$ as the glyph reconstruction.

\section{Learning and Inference}

Note that in our training setting, the font variables $Z$ are completely unobserved, and we must induce their manifold with learning.
As our model is generative, we wish to optimize the marginal probability of just the observed $X$ with respect to the model parameters $E$ and $\phi$:
\begin{equation*}
    p(X;E,\phi) = \int_{Z} p(X,Z;E,\phi)dZ
\end{equation*}
However, the integral over $Z$ is computationally intractable, given that the complex relationship between $Z$ and $X$ does not permit a closed form solution.
Related latent variable models such as Variational Autoencoders (VAE)~\citep{kingma2014auto} with intractable marginals have successfully performed learning by optimizing a variational lower bound on the log marginal likelihood. This surrogate objective, called the evidence lower bound (ELBO), introduces a variational approximation, $q(Z|X) = \prod_j q(z_i|x_{1j}, \ldots , x_{Ij})$ to the model's true posterior, $p(Z|X)$. Our model's ELBO is as follows:
\begin{equation*}
\begin{split}
    \mathrm{ELBO} =& \sum_{j} \mathbb{E}_q [ \log p(x_{1 j}, \ldots, x_{Ij}|z_j) ] \\
    & - \mathbb{KL}(q(z_j | x_{1j}, \ldots, x_{Ij}) || p(z_j))
\end{split}
\end{equation*}
where the approximation $q$ is parameterized via a neural encoder network.
This lower bound can be optimized by stochastic gradient ascent if $q$ is a Gaussian, via the reparameterization trick described in \cite{kingma2014auto, Rezende2014} to sample the expectation under $q$ while still permitting backpropagation.

Practically speaking, a key property which we desire is the ability to perform consistent inference over $z$ given a variable number of observed glyphs in a font. We address this in two ways: through the architecture of our encoder, and through a special masking process in training; both of which are shown in Figure~\ref{fig:inference}.

\begin{table*}
\centering
\resizebox{\textwidth}{!}{%
\begin{tabular}{@{}rcccc@{}c@{}ccccccccc@{}c@{}ccccc@{}}
\toprule
  & \multicolumn{4}{c}{Test Full} & \phantom{abc} & \multicolumn{4}{c}{Test Hard} & \phantom{abc} \\
\cmidrule{2-5} \cmidrule{7-10}
Observations  & 1 & 2 & 4 & 8 && 1 & 2 & 4 & 8  \\
\midrule
NN       & 483.13 & 424.49 & 386.81 & 363.97 && 880.22 & 814.67 & 761.29 & 735.18         \\
GlyphNet   & 669.36 & 533.97 & 455.23 & 416.65 && 935.01  & 813.50 & 718.02 & 653.57    \\
\midrule
Ours (FC)  & 353.63     & 316.47     & 293.67     & 281.89  && \textbf{596.57} & 556.21 & 527.50 & 513.25     \\
Ours (Conv)  & \textbf{352.07} & \textbf{300.46} & \textbf{271.03}  & \textbf{254.92}   && 615.87 & \textbf{556.03}  & \textbf{511.05}  & \textbf{489.58} \\
\bottomrule
\end{tabular}
}
\caption{$L_2$ reconstruction per glyph by number of observed characters. ``Full'' includes the entire test set while ``Hard'' is measured only over the 10\% of test fonts with the highest $L_2$ distance from the closest font in train.}
\label{tab:recon}
\end{table*}

\subsection{Posterior Approximation}

\textbf{Observation Subsampling:} To get reconstructions from only partially observed fonts at test time, the encoder network must be able to infer $z_j$ from any subset of $(x_{1j},\ldots,x_{Ij})$. One approach for achieving robustness to the number of observations is through the training procedure. Specifically when computing the approximate posterior for a particular font in our training corpus, we mask out a randomly selected subset of the characters before passing them to the encoder. This incentivizes the encoder to produce reasonable estimates without becoming too reliant on the features extracted from any one particular character, which more closely matches the setup at test time.

\noindent \textbf{Encoder Architecture:} Another way to encourage this robustness is through inductive bias in the encoder architecture. Specifically we use a convolutional neural network which takes in a batch of characters from a single font, concatenated with their respective character type embedding. Following the final convolutional layer, we perform an elementwise max operation across the batch, reducing to a single vector representation for the entire font which we pass through further fully-connected layers to obtain the output parameters of $q$ as shown in Figure~\ref{fig:inference}. By including this accumulation across the elements of the batch, we combine the features obtained from each character in a manner that is largely invariant to the total number and types of characters observed. This provides an inductive bias that encourages the model to extract similar features from each character type, which should therefore represent stylistic as opposed to structural properties.

Overall, the encoder over each glyph consists of three blocks, where each block consists of a convolution followed by a max pool with a stride of two, an instance norm~\citep{ulyanov2016instance}, and a ReLU. The activations are then pooled across the characters via an elementwise max into a single vector, which is then passed through four fully-connected layers, before predicting the parameters of the Gaussian approximate posterior.

\noindent \textbf{Reconstruction via Inference:} At test time, we pass an observed subset of a new font to our encoder in order to estimate the posterior over $z_j$, and take the mean of that distribution as the inferred font representation. We then pass this encoding to the decoder along with the full set of character embeddings $E$ in order to produce reconstructions of every glyph in the font.

\section{Experiments}

We now provide an overview of the specifics of the dataset and training procedure, and describe our experimental setup and baselines.

\subsection{Data}

We compare our model against baseline systems at font reconstruction on the Capitals64 dataset~\citep{GAN}, which contains the $26$ capital letters of the English alphabet as grayscale $64\times64$ pixel images across $10,682$ fonts. These are broken down into training, dev, and test splits of $7649$, $1473$, and $1560$ fonts respectively.

Upon manual inspection of the dataset, it is apparent that several fonts have an almost visually indistinguishable nearest neighbor, making the reconstruction task trivial using a naive algorithm (or an overfit model with high capacity) for those particular font archetypes. Because these datapoints are less informative with respect to a model's ability to generalize to previously \emph{unseen} styles, we additionally evaluate on a second test set designed to avoid this redundancy. Specifically, we choose the $10\%$ of test fonts that have maximal $L_2$ distance from their closest equivalent in the training set, which we call ``Test Hard''.

\subsection{Baselines}

As stated previously, many fonts fall into visually similar archetypes. Based on this property, we use a nearest neighbors algorithm for our first baseline. Given a partially observed font at test time, this approach simply ``reconstructs'' by searching the training set for the font with the lowest $L_2$ distance over the observed characters, and copy its glyphs verbatim for the missing characters.

For our second comparison, we use the GlyphNet model from~\citet{GAN}. This approach is based on a generative adversarial network, which uses discriminators to encourage the model to generate outputs that are difficult to distinguish from those in the training set. We test from the publicly available epoch $400$ checkpoint, with modifications to the evaluation script to match the setup described above.

We also perform an ablation using fully-connected instead of convolutional layers. For more architecture details see Appendix~\ref{sec:appendix}.

\subsection{Training Details}

We train our model to maximize the expected log likelihood using the Adam optimization algorithm~\citep{kingma2014adam} with a step size of $10^{-5}$ (default settings otherwise), and perform early stopping based on the approximate log likelihood on a hard subset of dev selected by the process described earlier. To encourage robustness in the encoder, we randomly drop out glyphs during training with a probability of $0.7$ (rejecting samples where all characters in a font are dropped). All experiments are run with a dimensionality of $32$ for the character embeddings and font latent variables. Our implementation\footnote{\url{https://bitbucket.org/NikitaSrivatsan/DeepFactorizationFontsEMNLP19}} is built in PyTorch~\cite{pytorch} version \texttt{1.1.0}.

\section{Results}

We now present quantitative results from our experiments in both automated and human annotated metrics, and offer qualitative analysis of reconstructions and the learned font manifold.

\subsection{Quantitative Evaluation}

\textbf{Automatic Evaluation:} We show font reconstruction results for our system against nearest neighbors and GlyphNet in Table~\ref{tab:recon}. Each model is given a random subsample of glyphs from each test font (we measure at $1$, $2$, $4$, and $8$ observed characters), with their character labels. We measure the average $L_2$ distance between the image reconstructions for the unobserved characters and the ground truth, after scaling intensities to $[0,1]$.

Our system achieves the best performance for both the overall and hard subset of test for all numbers of observed glyphs. Nearest neighbors provides a strong baseline on the full test set, even outperforming GlyphNet. However it performs much worse on the hard subset. This makes sense as we expect nearest neighbors to do extremely well on any test fonts that have a close equivalent in train, but suffer in fidelity on less traditional styles. GlyphNet similarly performs worse on test hard, which could reflect the \textit{missing modes} problem of GANs failing to capture the full diversity of the data distribution \cite{modes1,modes2}. The fully-connected ablation is also competitive, although we see that the convolutional architecture is better able to infer style from larger numbers of observations. On the hard test set, the fully-connected network even outperforms the convolutional system when only one observation is present, perhaps indicating that its lower-capacity architecture better generalizes from very limited data.

\newpage

\textbf{Human Evaluation:} To measure how consistent these perceptual differences are, we also perform a human evaluation of our model's reconstructions against GlyphNet using Amazon Mechanical Turk (AMT). In our setup, turkers were asked to compare the output of our model against the output of GlyphNet for a single font given one observed character, which they were also shown. Turkers selected a ranking based on which reconstructed font best matched the style of the observed character, and a separate ranking based on which was more realistic. On the full test set ($1560$ fonts, with $5$ turkers assigned to each font), humans preferred our system over GlyphNet $81.3\%$ and $81.8\%$ of the time for style and realism respectively. We found that on average $76\%$ of turkers shown a given pair of reconstructions selected the same ranking as each other on both criteria, suggesting high annotator agreement.

\begin{figure}
\includegraphics[width=0.49\textwidth]{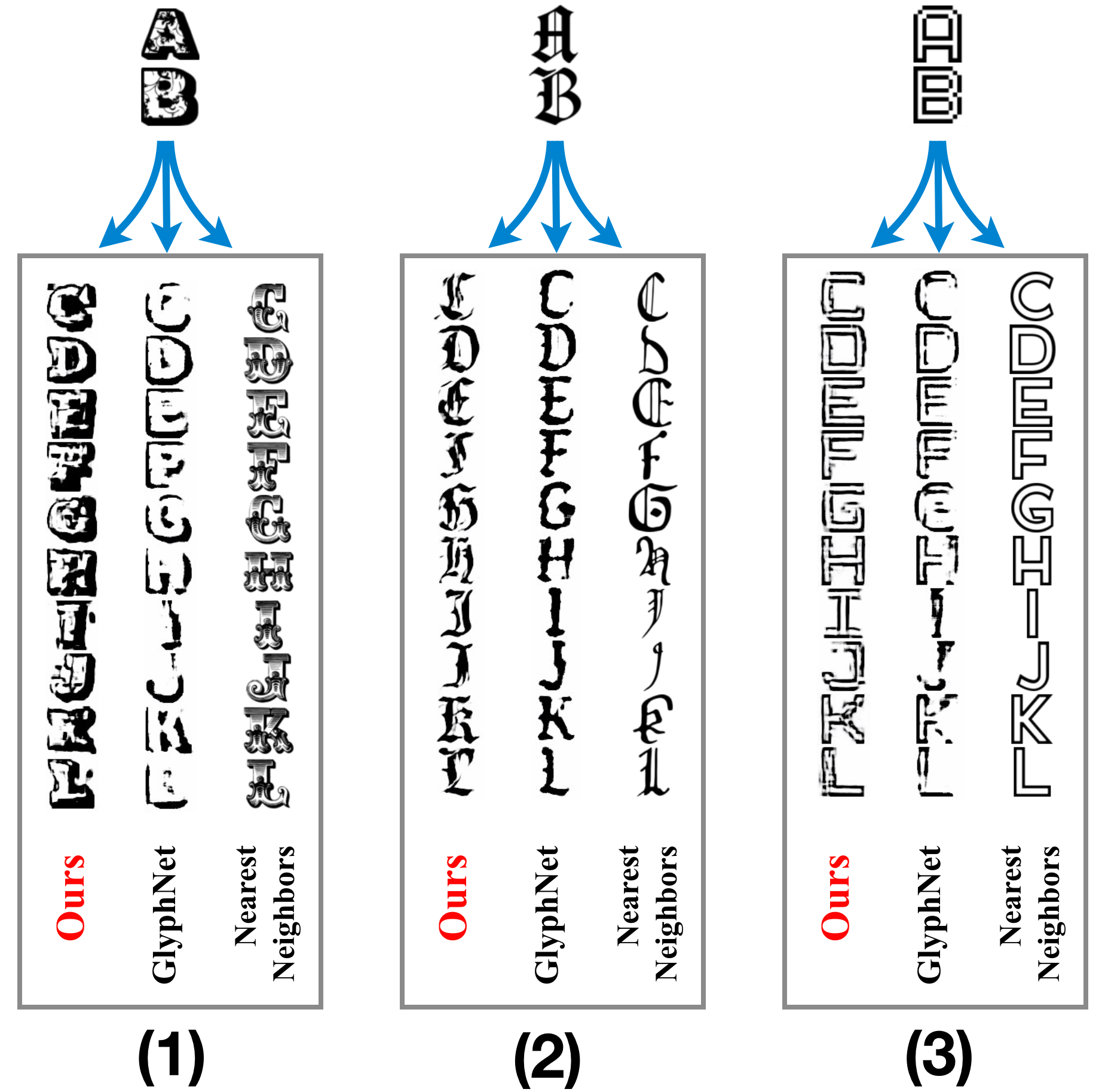}
\caption{Reconstructions of partially observed fonts in the hard subset from our model, GlyphNet, and nearest neighbors. Given images of glyphs for `A' and `B' in each font, we visualize reconstructions of the remaining characters. Fonts are chosen such that the $L_2$ loss of our model on these examples closely matches the average loss over the full evaluation set. \label{fig:complete}}
\end{figure}

\subsection{Qualitative Analysis}

In order to fully understand the comparative behavior of these systems, we also qualitatively compare the reconstruction output of these systems to analyze their various failure modes, showing examples in Figure~\ref{fig:complete}. We generally find that our approach tends to produce reconstructions that, while occasionally blurry at the edges, are generally faithful at reproducing the principal stylistic features of the font. For example, we see that for font (1) in Figure~\ref{fig:complete}, we match not only the overall shape of the letters, but also the drop shadow and to an extent the texture within the lettering, while GlyphNet does not produce fully enclosed letters or match the texture. The output of nearest neighbors, while well-formed, does not respect the style of the font as closely as it fails to find a font in training that matches these stylistic properties. In font (2) the systems all produce a form of gothic lettering, but the output of GlyphNet is again lacking in certain details, and nearest neighbors makes subtle but noticeable changes to the shape of the letters. In the final example (3) we even see that our system appears to attempt to replicate the pixelated outline, while nearest neighbors ignores this subtlety. GlyphNet is in this case somewhat inconsistent, doing reasonably well on some letters, but much worse on others. 
Overall, nearest neighbors will necessarily output well-formed glyphs, but with lower fidelity to the style, particularly on more unique fonts. While GlyphNet does pick up on some subtle features, our model tends to produce the most coherent output on harder fonts.

\begin{figure}
\scalebox{0.95}{\includegraphics[width=0.5\textwidth]{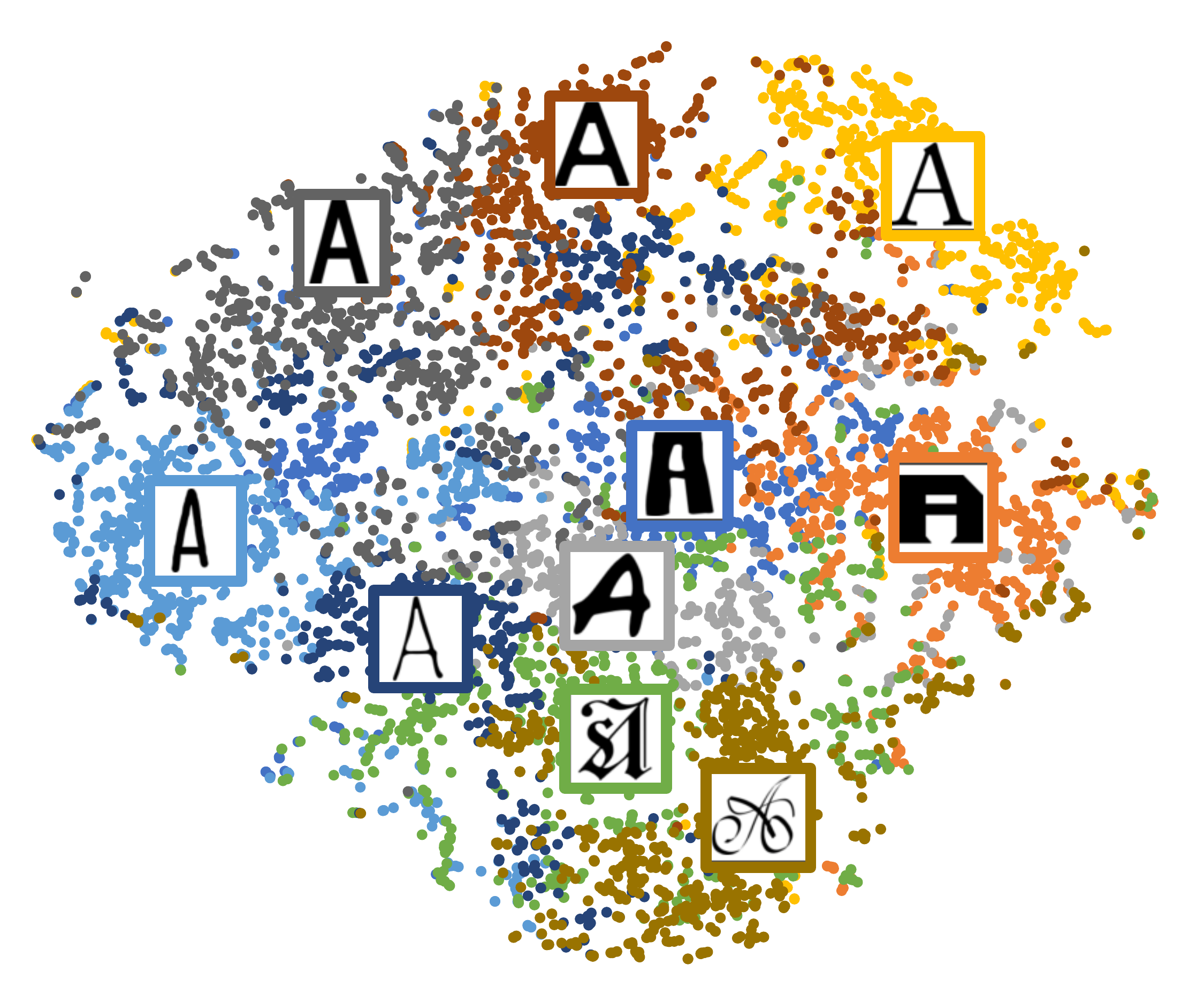}}
\caption{t-SNE projection of latent font variables inferred for the full training set, colored by k-means clustering with $k=10$. The glyph for the letter ``A'' for each centroid is shown in overlay. \label{fig:tsne}}
\end{figure}

\begin{figure*}

\begin{minipage}[c]{0.25\textwidth}
\caption{Interpolation between font variants from the same font family, showing smoothness of the latent manifold. Linear combinations of the embedded fonts correspond to outputs that lie intuitively ``in between''. \label{fig:interpolation}}
\end{minipage} \hfill
\begin{minipage}[c]{0.7\textwidth}
\includegraphics[width=\textwidth]{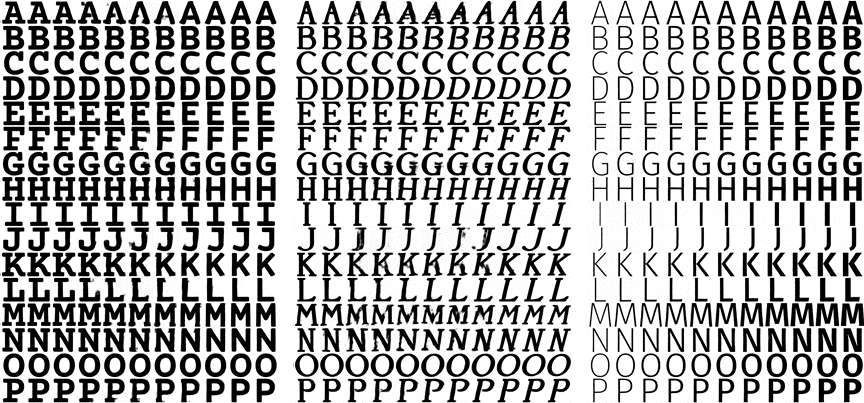}
\end{minipage}

\end{figure*}

\subsection{Analysis of Learned Manifold}

\begin{figure*}

\begin{minipage}[c]{0.46\textwidth}
\includegraphics[width=1\textwidth]{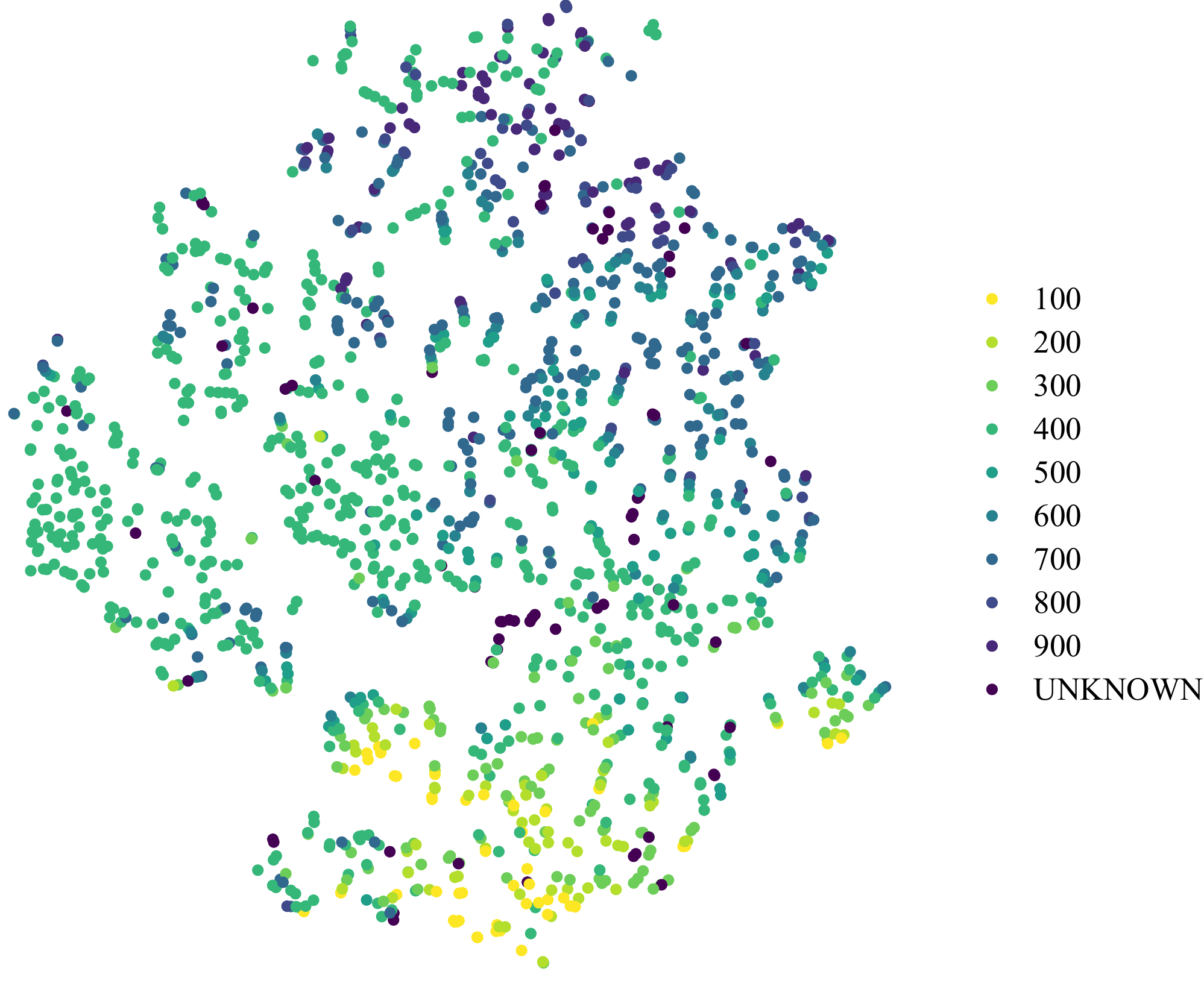}
\end{minipage} \hfill
\begin{minipage}[c]{0.46\textwidth}
\includegraphics[width=1\textwidth]{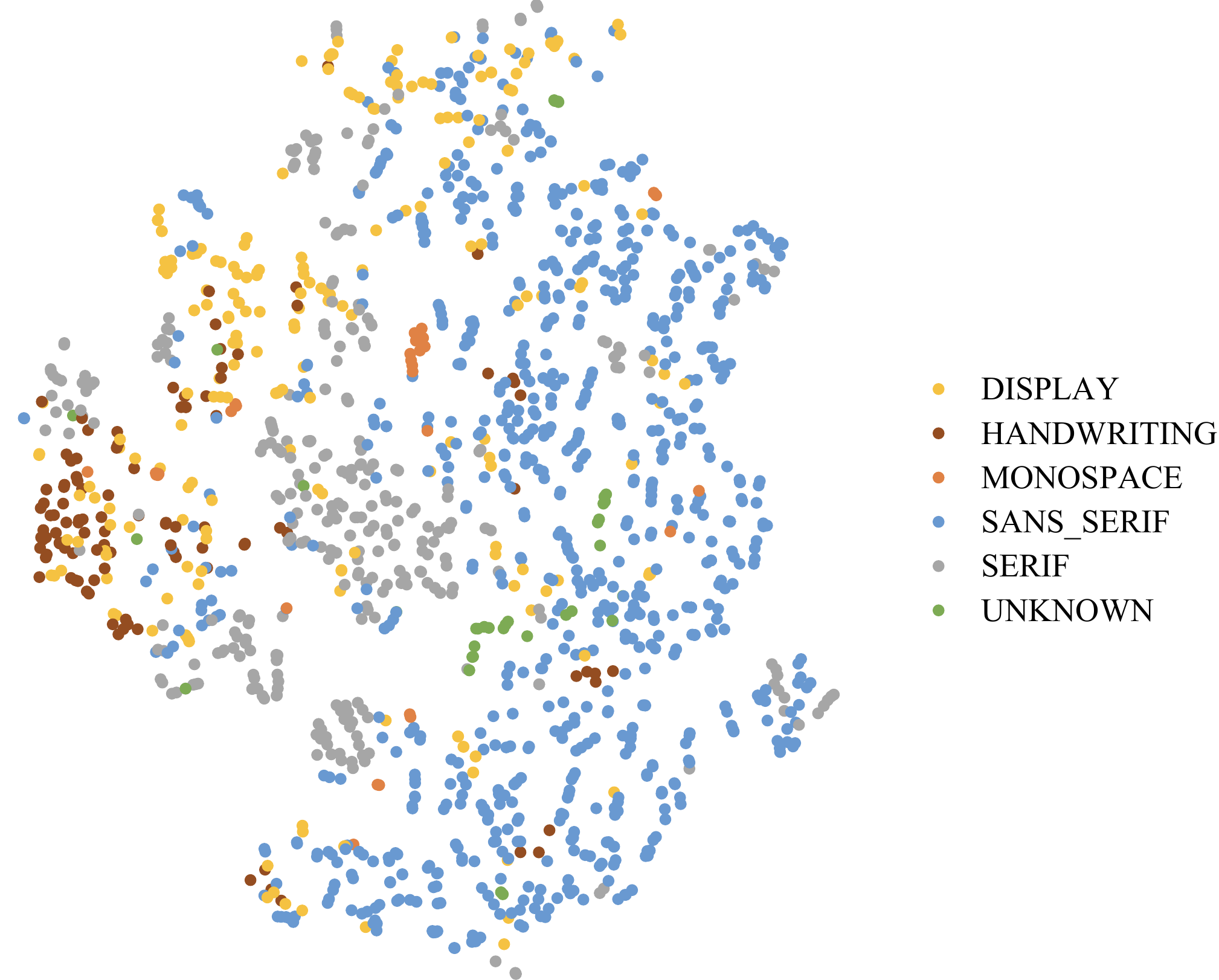}
\end{minipage} \hfill

\caption{t-SNE projection of latent font variables inferred on Google Fonts, colored by weight and category. 
\label{fig:tsne_goog}}
\end{figure*}

Since our model attempts to learn a smooth manifold over the latent style, we can also perform interpolation between the inferred font representations, something which is not directly possible using either of the baselines. In this analysis, we take two fonts from the same font family, which differ along one property, and pass them through our encoder to obtain the latent font variable for each. We then interpolate between these values, passing the result at various steps into our decoder to produce new fonts that exists in between the observations. In Figure~\ref{fig:interpolation} we see how our model can apply serifs, italicization, and boldness gradually while leaving the font unchanged in other respects. This demonstrates that our manifold is smooth and interpretable, not just at the points corresponding to those in our dataset. This could be leveraged to modify existing fonts with respect to a particular attribute to generate novel fonts efficiently.

Beyond looking at the quality of reconstructions, we also wish to analyze properties of the latent space learned by our model. To do this, we use our encoder to infer latent font variables $z$ for each of the fonts in our training data, and use a t-SNE projection~\citep{tsne} to plot them in 2D, shown in Figure~\ref{fig:tsne}. Since the broader, long-distance groupings may not be preserved by this transformation, we perform a k-means clustering~\citep{kmeans} with $k=10$ on the high-dimensional representations to visualize these high-level groupings. Additionally, we display a sample glyph for the centroid for each cluster. We see that while the clusters are not completely separated, the centroids generally correspond to common font styles, and are largely adjacent to or overlapping with those with similar stylistic properties; for example script, handwritten, and gothic styles are very close together.

To analyze how well our latent embeddings correspond to human defined notions of font style, we also run our system on the Google Fonts \footnote{\url{https://github.com/google/fonts}} dataset which despite containing fewer fonts and less diversity in style, lists metadata including numerical weight and category (e.g. \textit{serif}, \textit{handwriting}, \textit{monospace}). In Figure~\ref{fig:tsne_goog} we show t-SNE projections of latent font variables from our model trained on Google Fonts, colored accordingly. We see that the embeddings do generally cluster by weight as well as category suggesting that our model is learning latent information consistent with how humans perceive font style.

\section{Conclusion}

We presented a latent variable model of glyphs which learns disentangled representations of the structural properties of underlying characters from stylistic features of the font. We evaluated our model on the task of font reconstruction and showed that it outperformed both a strong nearest neighbors baseline and prior work based on GANs especially for fonts highly dissimilar to any instance in the training set. In future work, it may be worth extending this model to learn a latent manifold on content as well as style, which could allow for reconstruction of previously unseen character types, or generalization to other domains where the notion of content is higher dimensional.

\section*{Acknowledgments}
This project is funded in part by the NSF under grant 1618044 and by the NEH under grant HAA-256044-17.
Special thanks to Si Wu for help constructing the Google Fonts figures.

\bibliography{emnlp-ijcnlp-2019}
\bibliographystyle{acl_natbib}

\appendix

\section{Appendix}
\label{sec:appendix}

\subsection{Architecture Notation}

We now provide further details on the specific architectures used to parameterize our model and inference network. The following abbreviations are used to represent various components:
\begin{compactitem}
    \item $\text{F}_i$ : fully-connected layer with $i$ hidden units
    \item $\text{R}$ : ReLU activation
    \item $\text{M}$ : batch max pool
    \item $\text{S}$ : $2\times2$ spatial max pool
    \item $\text{C}_i$ : convolution filters with $i$ filters of $5\times5$, $2$ pixel zero-padding, stride of $1$, dilation of $1$
    \item $\text{I}$ : instance normalization 
    \item $\text{T}_{i,j,k}$ : transpose convolution with $i$ filters of $5\times5$, $2$ pixel zero-padding, stride of $j$, $k$ pixel output padding, dilation of $1$, where kernel and bias are the output of an MLP (described below)
    \item $\text{H}$ : reshape to $26\times256\times8\times8$
\end{compactitem}

\subsection{Network Architecture}

Our fully-connected encoder is: 

\noindent $\text{F}_{128}\text{-}\text{R}\text{-}\text{F}_{128}\text{-}\text{R}\text{-}\text{F}_{128}\text{-}\text{R}\text{-}\text{F}_{1024}\text{-}\text{R}\text{-}\text{M}\text{-}\text{F}_{128}\text{-}\text{R}\text{-}\text{F}_{128}\text{-}\\\text{R}\text{-}\text{F}_{128}\text{-}\text{R}\text{-}\text{F}_{64}$

\vspace{3mm}

\noindent
Our convolutional encoder is:

\noindent $\text{C}_{64}\text{-}\text{S}\text{-}\text{I}\text{-}\text{R}\text{-}\text{C}_{128}\text{-}\text{S}\text{-}\text{I}\text{-}\text{R}\text{-}\text{C}_{256}\text{-}\text{S}\text{-}\text{I}\text{-}\text{R}\text{-}\text{F}_{1024}\text{-}\text{M}\text{-}\text{R}\text{-}\\\text{F}_{128}\text{-}\text{R}\text{-}\text{F}_{128}\text{-}\text{R}\text{-}\text{F}_{128}\text{-}\text{R}\text{-}\text{F}_{64}$

\vspace{3mm}

\noindent
Our fully-connected decoder is:

\noindent $\text{F}_{128}\text{-}\text{R}\text{-}\text{F}_{128}\text{-}\text{R}\text{-}\text{F}_{128}\text{-}\text{R}\text{-}\text{F}_{128}\text{-}\text{R}\text{-}\text{F}_{128}\text{-}\text{R}\text{-}\text{F}_{4096}$

\vspace{3mm}

\noindent
Our transpose convolutional decoder is:

\noindent $\text{F}_{128}\text{-}\text{R}\text{-}\text{F}_{16384}\text{-}\text{R}\text{-}\text{H}\text{-}\text{T}_{256,2,1}\text{-}\text{R}\text{-}\text{C}_{256}\text{-}\text{I}\text{-}\text{R}\text{-}\text{C}_{256}\text{-}\text{I}\text{-}\\\text{R}\text{-}\text{T}_{128,2,1}\text{-}\text{R}\text{-}\text{C}_{128}\text{-}\text{I}\text{-}\text{R}\text{-}\text{C}_{128}\text{-}\text{I}\text{-}\text{R}\text{-}\text{T}_{64,2,1}\text{-}\text{I}\text{-}\text{R}\text{-}\text{C}_{64}\text{-}\\\text{I}\text{-}\text{R}\text{-}\text{C}_{64}\text{-}\text{I}\text{-}\text{R}\text{-}\text{T}_{32,1,0}\text{-}\text{I}\text{-}\text{R}\text{-}\text{C}_{32}\text{-}\text{I}\text{-}\text{R}\text{-}\text{C}_{1}$

\vspace{3mm}

\noindent
MLP to compute a transpose convolutional parameter of size $j$ is:

\noindent
$\text{F}_{128}\text{-}\text{R}\text{-}\text{F}_{j}$

\end{document}